\documentclass[twocolumn]{article}

\usepackage[utf8]{inputenc}
\usepackage[T1]{fontenc}
\usepackage{amsmath,amssymb}
\usepackage{graphicx}
\usepackage{booktabs}
\usepackage{multirow}
\usepackage{xcolor}
\usepackage[margin=1in]{geometry}
\usepackage{natbib}
\usepackage{hyperref}
\usepackage{microtype}
\usepackage{caption}
\usepackage{subcaption}
\usepackage{array}
\usepackage{tabularx}
\usepackage[most]{tcolorbox}

\newtcolorbox{pullquote}{
    colback=blue!3,
    colframe=blue!40!black,
    boxrule=1.5pt,
    arc=2pt,
    left=6pt, right=6pt, top=4pt, bottom=4pt,
    fontupper=\itshape\small,
    width=\columnwidth
}

\hypersetup{
    colorlinks=true,
    linkcolor=blue!60!black,
    citecolor=blue!60!black,
    urlcolor=blue!60!black
}

\title{\textbf{Semantic Novelty at Scale: Narrative Shape Taxonomy and Readership Prediction in 28,606 Books}}

\author{W.~Frederick Zimmerman\\
\textit{Nimble Books LLC}\\
\texttt{wfz@nimblebooks.com}}

\date{February 2026}

\begin{document}

\maketitle

\begin{figure*}[t]
    \centering
    \includegraphics[width=\textwidth]{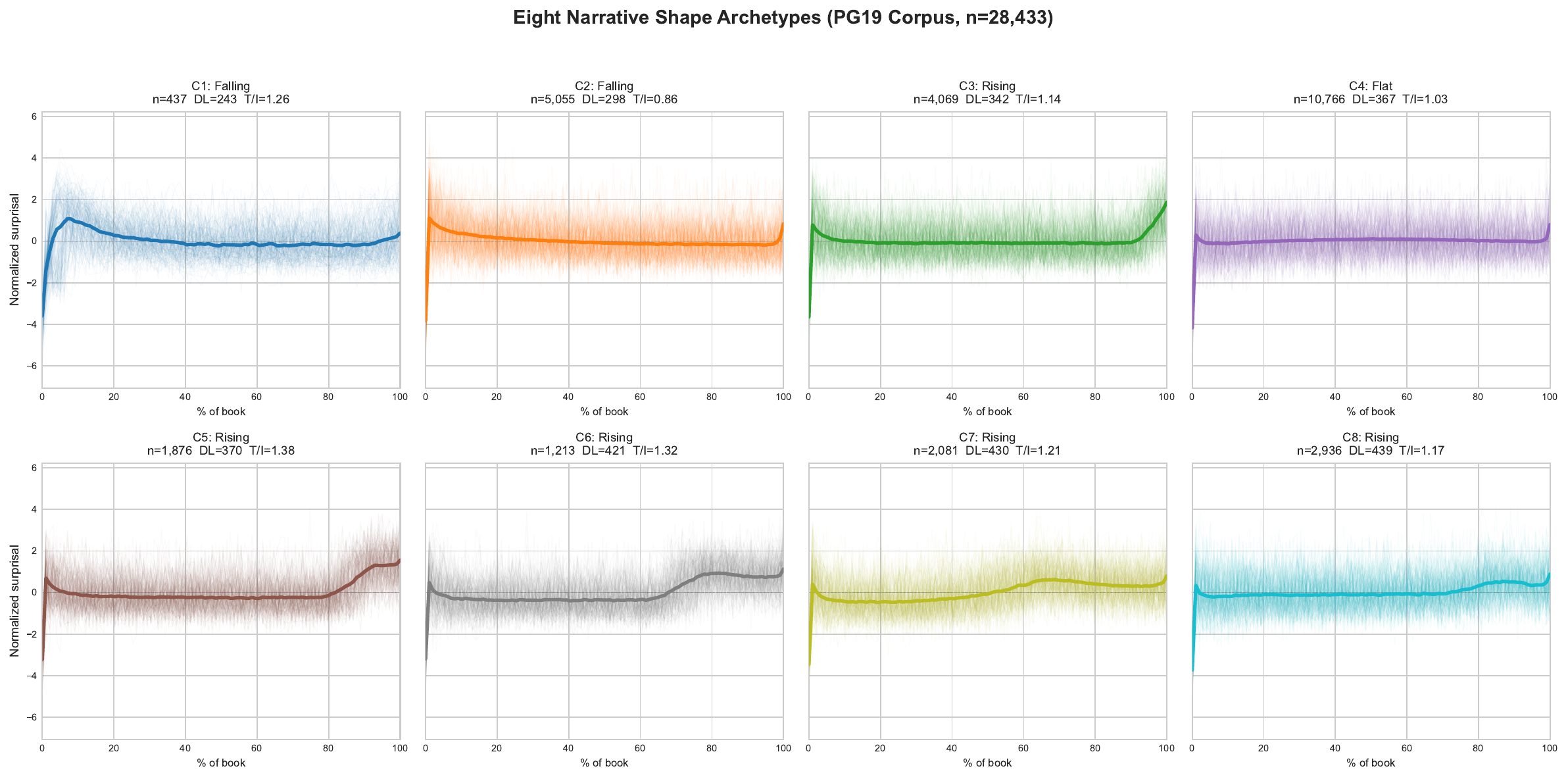}
    \caption{Eight narrative shape archetypes discovered by Ward-linkage hierarchical clustering on 16-segment PAA vectors from 28,606 books. Each panel shows the centroid curve (bold) with the 25th--75th percentile envelope (shaded). These eight shapes---from Steep Descent to Steep Ascent---constitute the core taxonomy of this paper.}
    \label{fig:hero_archetypes}
\end{figure*}


\begin{abstract}
We introduce \emph{semantic novelty}---the cosine distance between each paragraph's sentence embedding and the running centroid of all preceding paragraphs---as a new information-theoretic measure for characterizing narrative structure at corpus scale. Applying this measure to all 28,606 qualifying books in the PG19 corpus (pre-1920 English literature from Project Gutenberg), we compute paragraph-level novelty curves using 768-dimensional SBERT embeddings, then reduce each curve to a 16-segment Piecewise Aggregate Approximation (PAA) for systematic comparison. Ward-linkage hierarchical clustering on PAA vectors reveals \textbf{eight canonical narrative shape archetypes}, ranging from Steep Descent (rapid reader convergence) to Steep Ascent (escalating unpredictability). We find that \emph{volume}---the variance of the novelty trajectory, capturing how much semantic territory a book explores---is the strongest length-independent predictor of readership (partial $\rho = 0.32$, controlling for book length), followed by speed ($\rho = 0.19$) and Terminal/Initial ratio ($\rho = 0.19$). Circuitousness shows a strong raw correlation ($\rho = 0.41$) but is 93\% correlated with book length; after length control, its partial $\rho$ drops to 0.11. This methodological finding---that na\"ive correlations in corpus studies can be dominated by length confounds---carries implications beyond this study. Genre strongly constrains narrative shape ($\chi^2 = 2121.6$, $p < 10^{-242}$), with fiction maintaining plateau profiles while non-fiction front-loads information. Historical analysis reveals that books became progressively more predictable between 1840 and 1910 (T/I ratio trend $r = -0.74$, $p = 0.037$). Symbolic Aggregate Approximation (SAX) analysis shows 85\% signature uniqueness, suggesting each book traces a nearly unique path through semantic space. These findings demonstrate that information-density dynamics, distinct from sentiment or topic, constitute a fundamental dimension of narrative structure with measurable consequences for reader engagement. The full dataset is publicly available at \url{https://huggingface.co/datasets/wfzimmerman/pg19-semantic-novelty}.
\end{abstract}



\section{Introduction}
\label{sec:intro}

Reading is, at its core, a predictive act. Readers continually generate expectations about upcoming content based on what they have already encountered, and the interplay between expectation and surprise drives engagement \citep{schmidhuber2009}. When a paragraph closely matches the accumulated context of a book, it feels familiar; when it departs sharply, it introduces novelty. The dynamics of this tension---how information density evolves from opening to conclusion---constitute what we term the \emph{semantic novelty curve} of a text.

Prior work on narrative shape has predominantly relied on sentiment analysis. \citet{reagan2016} identified six emotional arcs in 1,327 Project Gutenberg novels using sentiment time series, while \citet{jockers2015} developed the Syuzhet package for extracting sentiment-based plot arcs. More recently, \citet{toubia2021} characterized narrative shape using topic model trajectories, defining metrics of speed, volume, and circuitousness for television narratives. \citet{boyd2020} proposed a narrative arc framework grounded in psychological processes. These approaches capture important dimensions of narrative---affect, topic, and psychological framing---but none directly measures the information-theoretic dimension: how surprising or predictable each passage is relative to the reader's accumulated model of the text.

\citet{schmidhuber2009} formalized this intuition in his Compression Progress theory, arguing that ``interestingness'' correlates with the rate at which an observer learns to compress (predict) incoming data. A passage that neither teaches the reader anything new (completely predictable) nor is entirely random (incompressible) is maximally interesting---it sits at the frontier of the reader's expanding model. Our semantic novelty measure operationalizes this framework at scale: for each paragraph, we compute the cosine distance between its embedding and the running centroid of all previous paragraph embeddings, yielding a continuous measure of how much each paragraph departs from the reader's accumulated context.

\begin{figure}[t]
    \centering
    \includegraphics[width=\columnwidth]{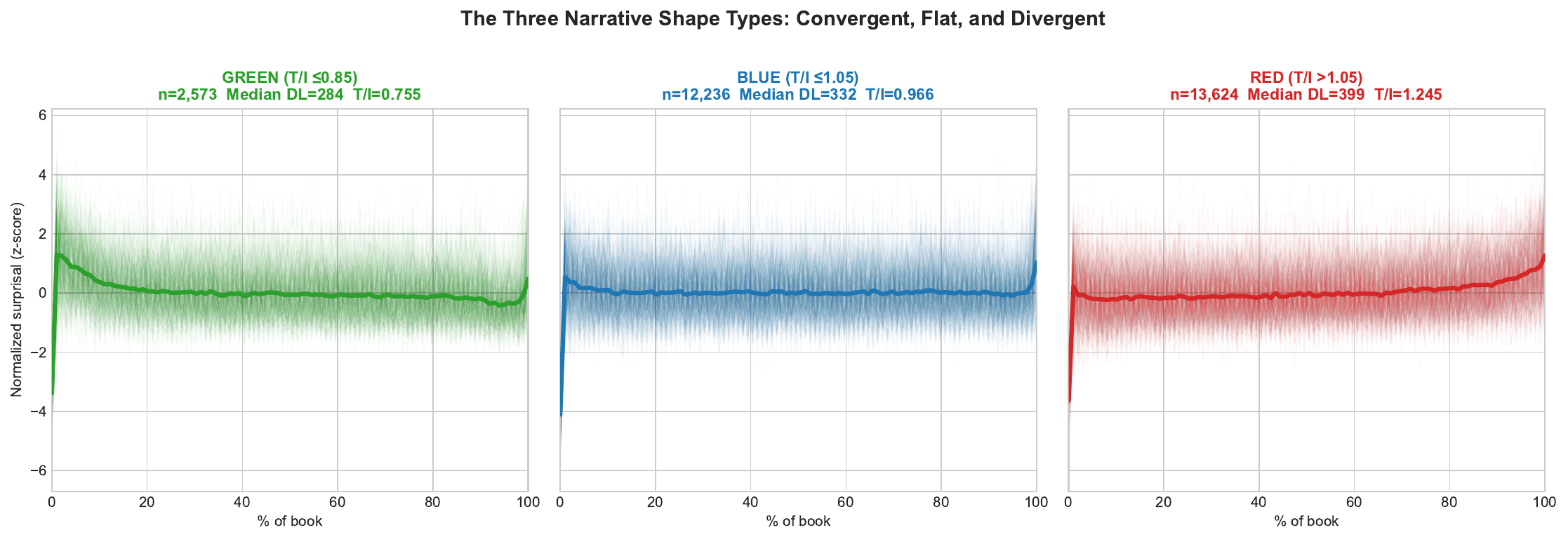}
    \caption{The three canonical novelty trajectory types: convergent (green, 22.5\%), plateau (blue, 59.0\%), and divergent (red, 18.6\%). These legacy categories are refined into eight archetypes by Ward-linkage clustering (Figure~\ref{fig:hero_archetypes}).}
    \label{fig:three_types}
\end{figure}

\begin{pullquote}
\textbf{Eight shapes of story.} Ward-linkage clustering reveals eight canonical narrative archetypes---from Steep Descent (rapid convergence) to Steep Ascent (escalating surprise)---providing the first fine-grained taxonomy of information-delivery shape at corpus scale.
\end{pullquote}

This paper makes five contributions:

\begin{enumerate}
    \item We introduce semantic novelty as a corpus-scale measure of information density, distinct from sentiment arcs and topic trajectories, and compute it for 28,606 books---an order of magnitude larger than previous narrative shape studies.
    \item We identify \textbf{eight canonical narrative shape archetypes} via Ward-linkage hierarchical clustering, providing a finer-grained taxonomy than previous 3--6 category systems.
    \item We demonstrate that \emph{volume}---how much semantic territory a text explores---is the strongest length-independent predictor of readership (partial $\rho = 0.32$), and that the apparently stronger raw correlation with circuitousness ($\rho = 0.41$) is largely a book-length confound, illustrating a methodological hazard for corpus-scale narrative studies.
    \item We show that genre conventions strongly constrain narrative shape ($p < 10^{-242}$), suggesting that genres encode implicit information-delivery contracts between authors and readers.
    \item We document a historical trend toward increasing predictability in English literature from 1840 to 1910.
\end{enumerate}

The remainder of this paper is organized as follows. Section~\ref{sec:related} reviews related work on narrative shape, compression progress, and time series methods. Section~\ref{sec:methods} describes our data, embedding procedure, and analytical pipeline. Section~\ref{sec:results} presents our findings across six subsections. Section~\ref{sec:discussion} interprets the results and addresses limitations. Section~\ref{sec:conclusion} summarizes contributions and outlines future directions.


\section{Related Work}
\label{sec:related}

\subsection{Narrative Shape and Sentiment Arcs}

The quantitative study of narrative shape emerged from the intersection of natural language processing and literary analysis. \citet{reagan2016} applied sentiment analysis to 1,327 Project Gutenberg novels, identifying six fundamental emotional arcs through singular value decomposition of sentiment time series: ``rags to riches,'' ``riches to rags,'' ``man in a hole,'' ``Icarus,'' ``Cinderella,'' and ``Oedipus.'' This work built on \citeauthor{vonnegut1995}'s (\citeyear{vonnegut1995}) informal taxonomy of story shapes. \citet{jockers2015} developed the Syuzhet R package for extracting sentiment-based plot arcs, using Fourier transforms to smooth sentiment trajectories. The approach generated productive debate about methodology \citep{swafford2015}, particularly regarding the sensitivity of results to smoothing parameters.

\citet{boyd2020} proposed a narrative arc framework grounded in three psychological dimensions---staging, plot progression, and cognitive tension---measured through linguistic markers rather than sentiment lexicons. Their approach demonstrates that narrative structure manifests in multiple linguistic dimensions simultaneously, supporting our argument that information density constitutes an additional, complementary dimension.

A fundamental limitation of sentiment-based approaches is that they capture affective valence but not informational content. A passage can be emotionally neutral yet informationally surprising (e.g., a technical revelation in a mystery novel), or emotionally intense yet informationally redundant (e.g., a repeated lament). Our semantic novelty measure addresses this gap.

\subsection{Narrative Shape via Topic Models}

\citet{toubia2021} introduced a productive framework for characterizing narrative shape using continuous metrics derived from topic model trajectories. Working with television narratives, they defined three metrics: \emph{speed} (mean absolute change between consecutive segments), \emph{volume} (variance of the trajectory), and \emph{circuitousness} (total path length divided by net displacement). They found that faster stories were more liked, while higher volume predicted lower ratings. We adopt their metric definitions but apply them to semantic novelty curves rather than topic trajectories, extending the framework from television to books and from topic space to embedding space.

\begin{figure}[t]
    \centering
    \includegraphics[width=\columnwidth]{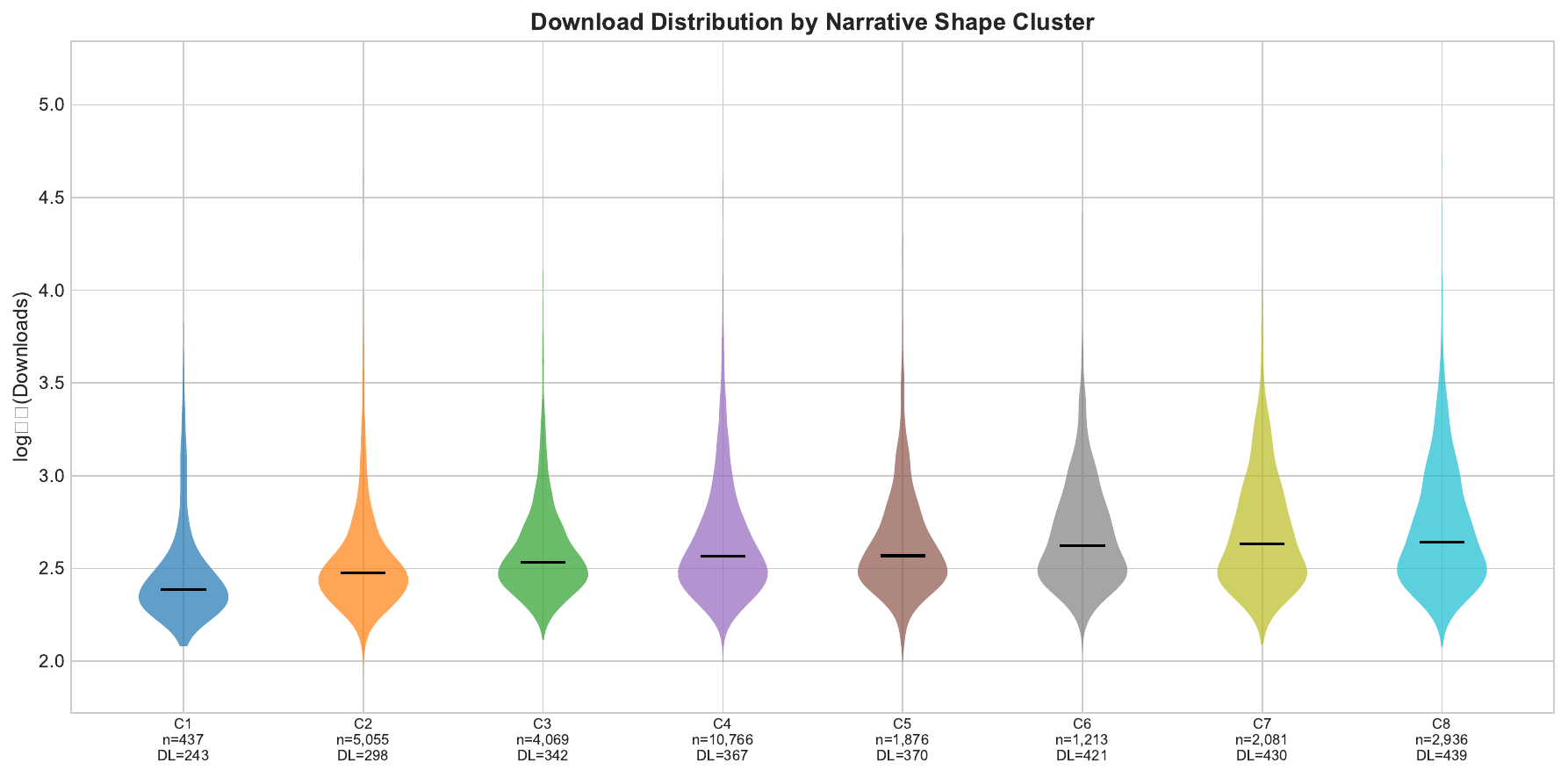}
    \caption{Distribution of $\log_{10}$(downloads) across the eight narrative clusters. Download distributions vary systematically by cluster, reflecting the interplay between narrative shape, book length, and readership patterns explored in Section~\ref{sec:results}.}
    \label{fig:download_cluster}
\end{figure}

\begin{pullquote}
\textbf{Semantic territory predicts readership.} After controlling for book length, \emph{volume}---the variance of the novelty curve---emerges as the strongest predictor of downloads (partial $\rho = 0.32$). Books that explore more semantic territory attract more readers, independent of how many pages they take.
\end{pullquote}

\subsection{Compression Progress and Information Theory}

\citet{schmidhuber2009} proposed that aesthetic appeal and ``interestingness'' arise from compression progress: the rate at which an observer's internal model improves its compression of incoming data. In this framework, completely predictable stimuli (zero compression progress) and completely random stimuli (no possible compression) are both uninteresting; maximal engagement occurs at the frontier where the observer is actively learning. Our semantic novelty measure can be interpreted through this lens: the first derivative of predictability (which we compute as mean compression progress) captures the rate at which the text becomes more or less predictable to a reader building an internal model.

\citet{karmarkar2024} investigated narrative reversals---directional changes in narrative trajectories---finding that reversals increase engagement in short-form narratives. We extend this analysis to book-length texts, computing reversal counts from smoothed semantic novelty curves.

\subsection{Time Series Representations}

Our analytical pipeline draws on established time series methods. Piecewise Aggregate Approximation \citep[PAA;][]{keogh2001} reduces variable-length time series to fixed-dimension vectors by averaging within equal-width segments, enabling efficient comparison. Symbolic Aggregate Approximation \citep[SAX;][]{lin2003sax} further discretizes PAA vectors into symbolic strings using breakpoints derived from the standard normal distribution, enabling pattern mining and motif discovery. Dynamic Time Warping \citep[DTW;][]{berndt1994dtw} computes similarity between time series that may vary in speed or alignment, finding the optimal warping path that minimizes total distance. Time series shapelets \citep{ye2009shapelet} identify discriminative subsequences that distinguish between classes, providing interpretable features for classification.

\subsection{Computational Literary Studies}

Our work contributes to the growing field of computational literary studies \citep{moretti2013, underwood2019, piper2018}. \citet{underwood2019} demonstrated that computational methods can reveal long-duration literary-historical trends invisible to close reading, examining changes in fiction across two centuries. \citet{piper2018} applied corpus-scale quantitative analysis to fundamental questions about literary form, arguing for ``enumerations'' as a mode of literary knowledge. \citet{moretti2013} pioneered ``distant reading'' as a complement to close reading, analyzing large collections to identify patterns. Our analysis of 28,606 books across three centuries continues this tradition, bringing information-theoretic tools to questions about narrative structure and literary change.


\section{Data and Methods}
\label{sec:methods}

\subsection{Corpus}

We use the PG19 corpus \citep{rao2023pg19}, comprising 28,752 pre-1920 English-language books from Project Gutenberg. After filtering for books with at least 20 paragraphs and valid novelty curves, our working corpus contains \textbf{28,606 books} spanning publication dates from the 1500s to 1920. The corpus covers diverse genres: Fiction (17.5\%), Children's/Juvenile (12.2\%), History (11.4\%), Social Science (7.2\%), Travel/Geography (5.6\%), Poetry (5.1\%), Biography (4.8\%), Philosophy/Religion (4.7\%), Science (3.9\%), Drama (2.5\%), and Other (25.1\%). The median book contains 288 paragraphs. Download counts from Project Gutenberg serve as a proxy for readership, ranging from 77 to 150,546 (median 356).

\subsection{Sentence Embeddings}

Each paragraph is encoded using SBERT \citep{reimers2019} with the \texttt{all-mpnet-base-v2} model, producing 768-dimensional dense vectors. This model achieves state-of-the-art performance on semantic textual similarity benchmarks and captures fine-grained semantic relationships beyond bag-of-words representations. Paragraphs are embedded independently, preserving the sequential nature of the text.

\subsection{Semantic Novelty Computation}

For each book, we compute paragraph-level semantic novelty as follows. Let $\mathbf{e}_i$ denote the embedding of paragraph $i$, and let $\mathbf{c}_i = \frac{1}{i-1}\sum_{j=1}^{i-1}\mathbf{e}_j$ denote the running centroid of all preceding paragraph embeddings. The semantic novelty of paragraph $i$ is:

\begin{equation}
    \text{novelty}_i = 1 - \frac{\mathbf{e}_i \cdot \mathbf{c}_i}{\|\mathbf{e}_i\| \, \|\mathbf{c}_i\|}
    \label{eq:novelty}
\end{equation}

This yields a value in $[0, 1]$, where 0 indicates perfect alignment with accumulated context and 1 indicates maximal departure. The running centroid accumulates the reader's ``model'' of the text, and novelty measures how much each paragraph challenges that model. Note that this is \emph{not} Shannon surprisal ($-\log p$); rather, it is a cosine-distance-based measure of semantic divergence from context. The running centroid models the reader as having perfect, equal-weight memory of all preceding content; in reality, human memory is recency-biased. We adopt the global centroid as a computationally tractable proxy for the cumulative ``world model'' a reader builds of the text, noting that an exponentially weighted alternative could capture recency effects.

We also compute several summary statistics: mean novelty, standard deviation of novelty, Terminal/Initial (T/I) ratio (mean novelty of last 10\% of paragraphs divided by mean novelty of first 10\%), trend slope (linear regression slope of novelty over paragraph index), and mean compression progress (mean of the first derivative of the negated novelty curve, capturing the rate of increasing predictability).

\subsection{Piecewise Aggregate Approximation and SAX}

To enable systematic comparison across books of varying length, each novelty curve is reduced to a 16-segment PAA vector. The curve is first z-normalized (mean 0, standard deviation 1), then divided into 16 equal-width segments, each represented by its mean value. This yields a fixed-length representation that preserves the overall shape while abstracting away length differences.

For symbolic analysis, PAA vectors are converted to SAX strings \citep{lin2003sax} using a 5-letter alphabet ($a$ through $e$) with breakpoints at the standard normal quantiles: $\{-0.84, -0.25, 0.25, 0.84\}$. Each segment's PAA value maps to one of five symbols, from $a$ (very low novelty relative to the book's own distribution) to $e$ (very high). The resulting 16-character strings provide a compact symbolic signature of each book's novelty trajectory.

\subsection{Clustering}

We apply Ward-linkage hierarchical clustering \citep{ward1963} to z-scored 16-segment PAA vectors for a random sample of 8,000 books, identifying eight clusters that capture distinct narrative shape archetypes. All 28,606 books are then assigned to the nearest cluster centroid by Euclidean distance. We selected $k=8$ based on multiple criteria: silhouette scores (which peak at $k=2$ at 0.091 and decline monotonically, reflecting the continuous nature of literary data rather than discrete clusters), WCSS elbow analysis (which shows diminishing returns in variance reduction beyond $k=6$, with marginal reduction dropping from 3.4\% at $k=5$ to 1.8\% at $k=8$ to 1.1\% at $k=10$; see Table~\ref{tab:wcss} in Appendix~\ref{app:kselection}), dendrogram inspection, and interpretability. The low silhouette scores across all $k$ reflect that narrative shapes form a continuum rather than discrete categories; our clusters are best understood as prototypical regions in a continuous shape space. For comparison, we also compute DTW-based clustering on a subsample of 2,000 books.

\subsection{Toubia Shape Metrics}

Following \citet{toubia2021}, we compute three continuous shape metrics from the raw novelty curves:

\begin{align}
    \text{Speed} &= \frac{1}{n-1} \sum_{i=1}^{n-1} |\text{novelty}_{i+1} - \text{novelty}_i| \\
    \text{Volume} &= \text{Var}(\text{novelty}_1, \ldots, \text{novelty}_n) \\
    \text{Circuitousness} &= \frac{\sum_{i=1}^{n-1}|\text{novelty}_{i+1} - \text{novelty}_i|}{|\text{novelty}_n - \text{novelty}_1|}
\end{align}

Speed captures the average rate of change; volume captures the overall spread; circuitousness captures how winding the path is relative to net displacement. Note that speed is length-normalized by definition (mean over $n-1$ steps), and volume depends on distributional spread rather than series length. However, circuitousness accumulates absolute changes over all paragraphs, making its numerator mechanically proportional to text length; this creates a critical confound that we address in Section~\ref{sec:confound}. We additionally compute acceleration (mean absolute second derivative), roughness (mean absolute third derivative), peak count, and curve entropy.

\subsection{Reversal Count}

Following \citet{karmarkar2024}, we count direction reversals in the smoothed novelty curve. The curve is smoothed with a rolling mean (window = 10 paragraphs), the first derivative is computed, and reversals are counted as sign changes in the derivative. This captures how many times the narrative shifts between increasing and decreasing novelty.

\subsection{Genre Classification}

Books are classified into 11 genre categories using regex-based matching against Library of Congress subject headings provided in the PG19 metadata. Categories are: Fiction, Poetry, Drama, History, Science, Philosophy/Religion, Travel/Geography, Biography, Children's/Juvenile, Social Science, and Other. The classification is rule-based and approximate, prioritizing recall over precision.


\section{Results}
\label{sec:results}

\subsection{Eight-Cluster Narrative Taxonomy}

Ward-linkage hierarchical clustering on 8,000 randomly sampled PAA vectors, followed by nearest-centroid assignment of all 28,606 books, yields eight interpretable narrative shape archetypes (Figure~\ref{fig:hero_archetypes}):

\begin{enumerate}
    \item \textbf{Steep Descent} (5.9\%): Novelty drops sharply from start to finish, indicating rapid reader convergence with the material.
    \item \textbf{Gradual Descent} (0.8\%): Moderate, steady decrease in novelty throughout.
    \item \textbf{Early Plateau} (15.8\%): Quick convergence in opening sections, then stable novelty for the remainder.
    \item \textbf{Late Plateau} (23.0\%): Novelty stays elevated initially, then converges late.
    \item \textbf{U-Shape} (9.8\%): Novelty decreases in the first half, then rises again.
    \item \textbf{Flat} (26.2\%): Minimal variation throughout---the most common profile.
    \item \textbf{Gradual Ascent} (9.3\%): Slowly increasing novelty from beginning to end.
    \item \textbf{Steep Ascent} (9.3\%): Rapidly escalating novelty, the text becoming increasingly surprising.
\end{enumerate}

These eight clusters refine the legacy three-type classification (convergent, plateau, divergent) that corresponds to trend direction alone. Convergent curves (clusters 1--3) constitute 22.5\% of the corpus; plateau/mixed curves (clusters 4--6) constitute 59.0\%; and divergent curves (clusters 7--8) constitute 18.6\%.

The dominance of Flat (26.2\%) and Late Plateau (23.0\%) profiles indicates that nearly half of all books maintain relatively stable novelty levels---the running centroid quickly captures the semantic character of the text, and subsequent paragraphs offer consistent, moderate departure from it.


\subsection{Genre-Curve Association}

Genre and narrative shape are strongly associated. A chi-square test of genre $\times$ curve type (the three-way convergent/plateau/divergent classification) yields $\chi^2 = 2121.6$ (dof~=~20, $p < 10^{-100}$), and a Kruskal-Wallis test of genre $\to$ T/I ratio yields $H = 1158.7$ ($p = 1.15 \times 10^{-242}$).

\begin{pullquote}
\textbf{Genre as information contract.} Genre conventions constrain narrative shape with extraordinary strength ($\chi^2 = 2121.6$, $p < 10^{-242}$): fiction maintains plateau profiles while non-fiction front-loads information, suggesting genres encode implicit expectations about how ideas will be delivered.
\end{pullquote}

Table~\ref{tab:genre} summarizes key metrics by genre. Fiction shows the most distinctive profile: a T/I ratio of 1.022 (near-unity, indicating balanced novelty) and a strong plateau preference (65\% blue). Note that circuitousness values in the table are strongly confounded with typical book length per genre (Section~\ref{sec:confound}); fiction's high circuitousness (264.9) partly reflects that novels are longer than poems or pamphlets. Non-fiction texts, by contrast, have higher T/I ratios (mean 1.077), indicating front-loaded information delivery. This difference is highly significant (Mann-Whitney $U = 3.37 \times 10^7$, $p = 5.72 \times 10^{-73}$).

Travel/Geography shows the highest speed (0.131), reflecting rapid topic shifts as narratives move through geographical descriptions. Biography has the highest median downloads (485). Poetry has the highest percentage of convergent curves (20\%), consistent with the concentrated, distilled nature of poetic language.

\begin{table}[t]
\centering
\caption{Semantic novelty metrics by genre. T/I = Terminal/Initial ratio; Circ = median circuitousness; DL = median downloads.}
\label{tab:genre}
\small
\begin{tabular}{@{}lrrrr@{}}
\toprule
Genre & $N$ & T/I & Circ & Med DL \\
\midrule
Fiction & 4,974 & 1.022 & 264.9 & 315 \\
Children's & 3,470 & 1.072 & 189.4 & 291 \\
History & 3,242 & 1.088 & 210.6 & 424 \\
Social Science & 2,042 & 1.059 & 199.2 & 353 \\
Travel/Geogr. & 1,598 & 1.118 & 157.7 & 474 \\
Poetry & 1,449 & 1.031 & 69.4 & 319 \\
Biography & 1,363 & 1.075 & 175.1 & 485 \\
Phil./Religion & 1,333 & 1.079 & 142.8 & 373 \\
Science & 1,095 & 1.128 & 159.4 & 414 \\
Drama & 719 & 1.024 & 180.7 & 312 \\
Other & 7,148 & 1.129 & 117.6 & 381 \\
\bottomrule
\end{tabular}
\end{table}

\begin{figure}[t]
    \centering
    \includegraphics[width=\columnwidth]{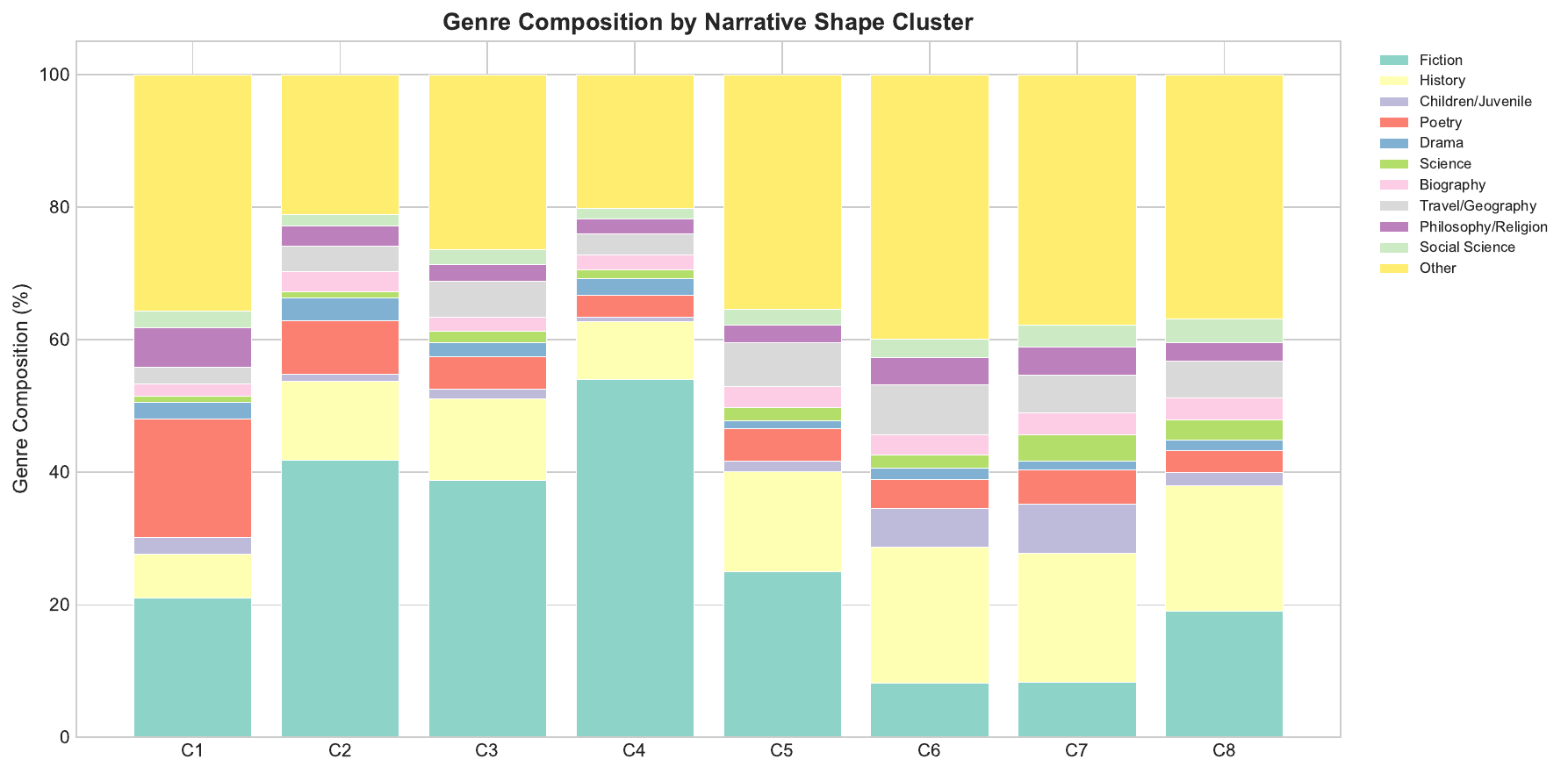}
    \caption{Genre composition within each of the eight narrative clusters. Fiction dominates plateau clusters (Flat, Late Plateau); History and Other dominate descent and ascent clusters. Genre and cluster are strongly associated ($\chi^2 = 543.2$, $p = 2.41 \times 10^{-74}$).}
    \label{fig:genre_cluster}
\end{figure}

\subsection{Toubia Metrics and the Length Confound}
\label{sec:confound}

Table~\ref{tab:toubia} presents Spearman correlations between shape metrics and $\log_{10}$(downloads), both raw and after controlling for book length (paragraph count) via partial Spearman correlation. The results reveal a critical methodological finding: \textbf{the raw rankings are dominated by length confounds.}

In the raw correlations, circuitousness appears to be the strongest predictor ($\rho = 0.406$), followed by compression progress ($\rho = 0.388$) and reversal count ($\rho = 0.381$). However, all three of these metrics correlate with book length at $\rho > 0.92$. Circuitousness accumulates absolute changes across all paragraphs; compression progress and reversal count similarly grow with text length. Since longer books also tend to have more downloads ($\rho_{\text{length,downloads}} = 0.37$), the apparently strong prediction is substantially a proxy for ``longer books are downloaded more.''

After partialling out paragraph count, the ranking reshuffles completely. \textbf{Volume}---the variance of the novelty trajectory, measuring how much semantic territory a book explores---emerges as the strongest length-independent predictor (partial $\rho = 0.32$). Standard deviation of novelty, a closely related measure, shows the same strength ($\rho = 0.32$). These are followed by T/I ratio ($\rho = 0.19$), speed ($\rho = 0.19$), and trend slope ($\rho = 0.17$). Circuitousness retains a modest partial $\rho = 0.11$---still significant given $n = 28{,}433$, but no longer dominant.

\begin{table}[t]
\centering
\caption{Spearman correlations with $\log_{10}$(downloads): raw vs.\ length-controlled, $n = 28{,}433$. Metrics marked $\dagger$ are strongly correlated with book length ($|\rho_{\text{length}}| > 0.90$).}
\label{tab:toubia}
\small
\begin{tabular}{@{}lrrr@{}}
\toprule
Metric & $\rho_{\text{raw}}$ & $\rho_{\text{partial}}$ & $\rho_{\text{length}}$ \\
\midrule
Circuitousness$^\dagger$ & 0.406 & 0.113 & 0.928 \\
Compr.\ progress$^\dagger$ & 0.388 & 0.040 & 0.944 \\
Reversal count$^\dagger$ & 0.381 & $-0.180$ & 0.998 \\
\addlinespace
\textbf{Volume} & \textbf{0.049} & \textbf{0.317} & $-0.495$ \\
Std novelty & 0.046 & 0.316 & $-0.500$ \\
T/I ratio & 0.207 & 0.191 & 0.088 \\
Speed & 0.005 & 0.187 & $-0.375$ \\
Trend slope & 0.130 & 0.170 & $-0.009$ \\
Mean novelty & $-0.028$ & $-0.123$ & 0.206 \\
\bottomrule
\end{tabular}
\end{table}

Our results both replicate and revise \citet{toubia2021}. Speed shows a non-significant raw correlation ($\rho = 0.005$, $p = 0.37$), but after length control reveals a meaningful positive relationship (partial $\rho = 0.19$), now consistent with Toubia's finding that faster narratives are preferred. The reversal of the volume effect is more dramatic: \citeauthor{toubia2021} found that higher volume predicted \emph{lower} ratings for television; we find it predicts \emph{higher} downloads for books. This divergence likely reflects fundamental differences between media: in a 30-minute episode, high semantic variance can feel chaotic, but in a book-length text, sustained exploration of diverse semantic territory reads as richness rather than incoherence.

The length confound carries a methodological lesson for the field: summative metrics computed over variable-length sequences (total path length, reversal count, compression progress) will correlate mechanically with length. Studies of narrative shape should report length-controlled partial correlations alongside raw values, or use length-normalized variants.

\subsection{Volume, Semantic Territory, and Readership}
\label{sec:volume}

The volume--downloads relationship merits detailed examination. Volume (the variance of the novelty curve) captures how broadly a text ranges across semantic space---high-volume books traverse diverse topics and registers, while low-volume books remain in a narrow semantic band. After controlling for length, volume is the strongest predictor of readership (partial $\rho = 0.32$, $p < 10^{-300}$, $n = 28{,}433$).

An OLS regression of $\log_{10}$(downloads) on log-circuitousness, log-paragraph-count, publication year, speed, volume, mean novelty, T/I ratio, and genre indicators ($R^2 = 0.18$, $n = 8{,}996$ with complete data) confirms this finding in a multivariate framework: volume is the strongest standardized predictor ($\beta = 0.053$, $t = 9.3$), followed by log-circuitousness ($\beta = 0.120$, $t = 11.4$). Variance inflation factors confirm the collinearity concern: $\text{VIF}_{\log(\text{circ})} = 10.3$ and $\text{VIF}_{\log(\text{para})} = 11.7$ both exceed the conventional threshold of 10, while $\text{VIF}_{\text{volume}} = 3.0$, $\text{VIF}_{\text{speed}} = 2.5$, and $\text{VIF}_{\text{T/I}} = 1.4$ remain well-behaved. When circuitousness is removed from the model, $R^2$ drops by only 0.012; when paragraph count is removed instead, $R^2$ drops by merely 0.001---confirming that circuitousness contributes little beyond what book length already explains.

\textbf{Robustness.} The volume--readership relationship holds within major genres: partial $\rho = 0.19$ within Fiction ($n = 11{,}075$, $p < 10^{-93}$), $0.18$ within Drama ($n = 688$, $p < 10^{-6}$), and $0.12$ within Philosophy/Religion ($n = 729$, $p = 0.001$). It is weaker or absent in reference-oriented genres (History: $0.06$; Science: $-0.07$; Travel: $-0.02$), where downloads likely reflect information-seeking rather than narrative engagement. The effect also holds within decade bins: partial $\rho = 0.35$ for 1860--1879, $0.39$ for 1880--1899, and $0.29$ for 1900--1920 (all $p < 10^{-4}$), ruling out era-driven confounds. These within-genre and within-decade results confirm that volume's predictive power is not an artifact of genre composition or historical period.

The circuitousness quintile analysis (Table~\ref{tab:circ_quintile}) illustrates this confound clearly: the Q1-to-Q5 doubling in median downloads (267 to 539) parallels a corresponding increase in median paragraph count. Nevertheless, the curve-type moderation effect is noteworthy: within divergent (red) curves, circuitousness correlates with downloads at $\rho = 0.497$, higher than within plateau ($\rho = 0.331$) or convergent ($\rho = 0.305$) curves. Even after length control, some residual circuitousness signal persists (partial $\rho = 0.11$), suggesting that path complexity contributes modestly to engagement beyond mere length.

A Kruskal-Wallis test confirms that the three-way curve type classification significantly predicts downloads: $H = 1148.8$, $p = 3.52 \times 10^{-250}$. Red (divergent) books have higher median downloads (399) than blue (332) or green (284).

\begin{table}[t]
\centering
\caption{Downloads by circuitousness quintile. The gradient parallels book length; see Section~\ref{sec:confound} for length-controlled analysis.}
\label{tab:circ_quintile}
\small
\begin{tabular}{@{}lrrrr@{}}
\toprule
Quintile & Med Circ & $N$ & Med DL & \%Red \\
\midrule
Q1 (straight) & 37.0 & 5,687 & 267 & 44.8\% \\
Q2 & 97.4 & 5,686 & 329 & 52.1\% \\
Q3 & 172.1 & 5,687 & 368 & 52.7\% \\
Q4 & 275.4 & 5,686 & 402 & 47.6\% \\
Q5 (winding) & 506.9 & 5,687 & 539 & 42.4\% \\
\bottomrule
\end{tabular}
\end{table}

\begin{figure}[t]
    \centering
    \includegraphics[width=\columnwidth]{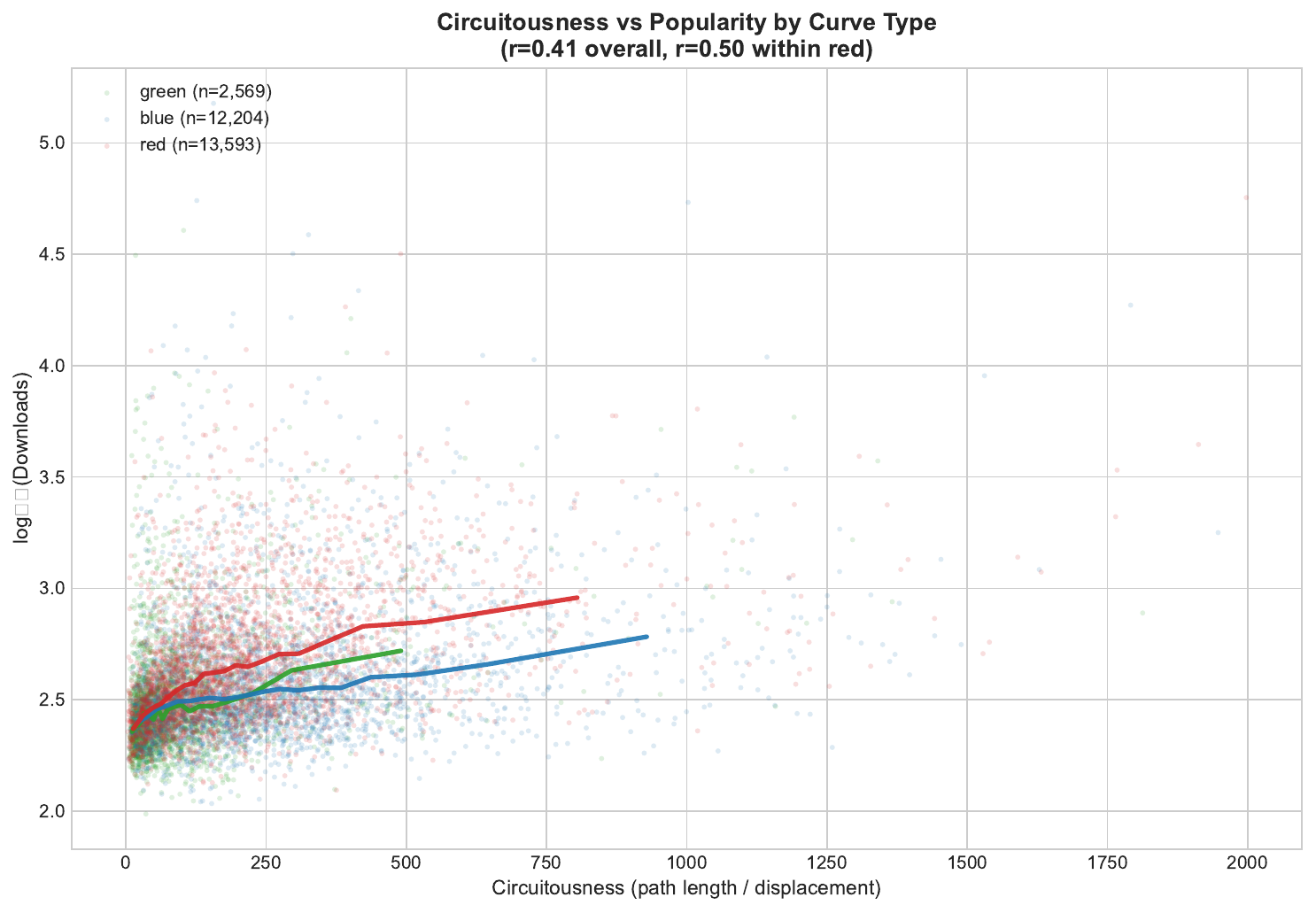}
    \caption{Circuitousness vs.\ $\log_{10}$(downloads), colored by curve type. The raw Spearman $\rho = 0.41$ is substantially confounded by book length ($\rho_{\text{circ,length}} = 0.93$); after length control, the partial $\rho = 0.11$. See Table~\ref{tab:toubia} for length-controlled rankings.}
    \label{fig:circ_scatter}
\end{figure}

\subsection{Shapelets and SAX Patterns}

SAX analysis reveals that semantic novelty trajectories are highly individualized. From 28,433 books, we obtain 24,221 unique 16-character SAX signatures---an 85.2\% uniqueness ratio. The most common signature, \texttt{cccccccccccccccc} (consistently mid-level novelty), occurs in only 571 books (2.0\%). This high uniqueness suggests that while books cluster into broad shape families, each traces a nearly unique path through semantic space at the symbolic level.

\begin{figure}[t]
    \centering
    \includegraphics[width=\columnwidth]{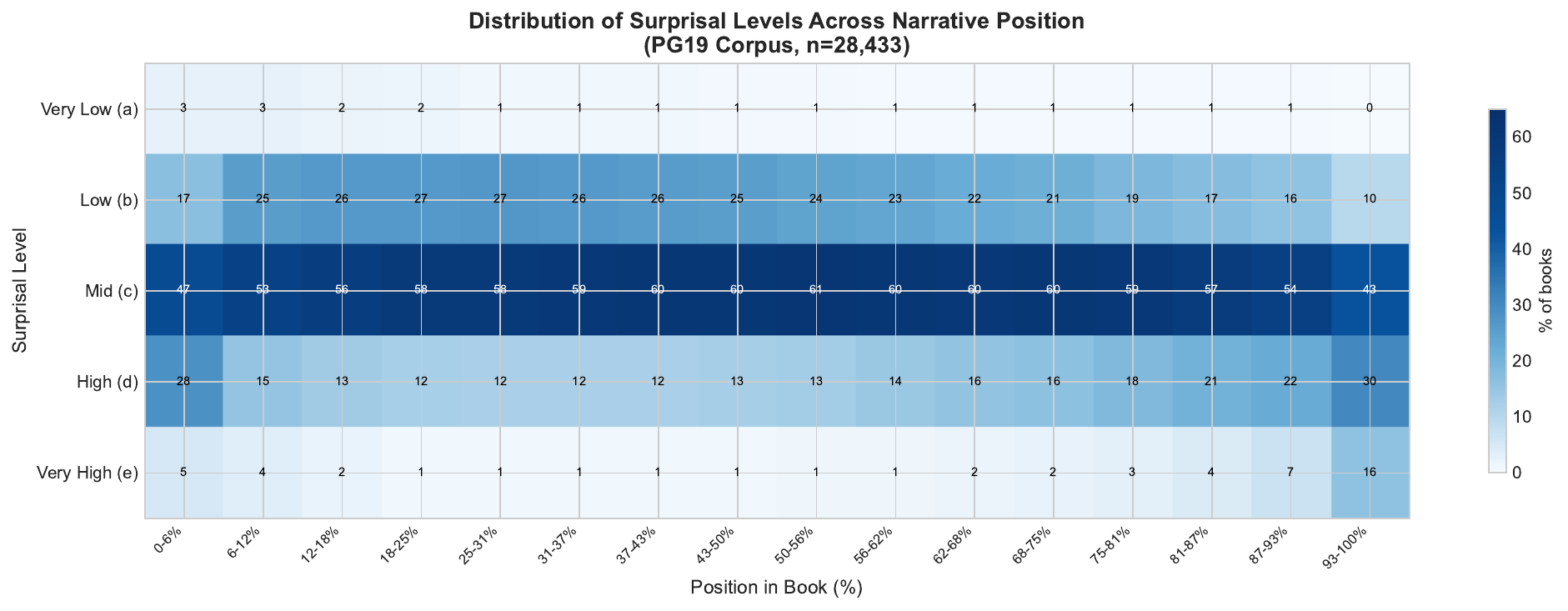}
    \caption{SAX symbol frequency by position. The opening and closing positions show the greatest variance; middle positions converge toward $c$ (mid-level novelty). The rightward shift from $b$/$c$ dominance to $c$/$d$ dominance reflects the overall tendency toward slightly increasing novelty.}
    \label{fig:sax}
\end{figure}

Position-by-position analysis of SAX symbols (Figure~\ref{fig:sax}) reveals systematic structure. The opening position shows high variance (5.1\% $e$, 28.4\% $d$), reflecting the diversity of opening strategies. Middle positions converge toward $c$ (mid-level, peaking at 60.1\% in position 8). The final position shows the widest spread (16.2\% $e$, 30.2\% $d$), consistent with the diversity of endings. This pattern---high variance at boundaries, stability in the middle---recapitulates the well-known observation that beginnings and endings carry disproportionate structural weight.

Shapelet discovery identifies discriminative curve fragments. For genre discrimination (fiction vs.\ non-fiction), the top shapelets achieve information gains of 0.033--0.044 and concentrate in opening (32\%) and middle (30\%) positions. For popularity discrimination (high vs.\ low downloads), shapelets concentrate in middle (34\%) and late (26\%) positions, suggesting that the middle and latter portions of a book's novelty trajectory are most predictive of readership. Cross-genre analysis reveals that Poetry vs.\ Drama is most distinguishable by shapelets (best IG = 0.085), followed by Fiction vs.\ History (IG = 0.044).

\subsection{Historical Trends}

Publication era analysis reveals a systematic shift in narrative shape over the period 1840--1910 (Table~\ref{tab:era}). The mean T/I ratio decreases from 1.078 in the 1840s to 1.024 in the 1910s ($r = -0.74$, $p = 0.037$), indicating that books became progressively more predictable---their endings deviated less from their beginnings in terms of novelty. Concurrently, the proportion of divergent (red) curves dropped from 51.7\% to 33.5\%, while plateau (blue) curves rose from 24.1\% to 58.1\%.

This trend coincides with the industrialization of publishing, the rise of lending libraries (particularly Mudie's Circulating Library), and increasing editorial professionalization. These market forces may have selected for more conventionally structured narratives---texts that delivered information in predictable patterns rather than escalating novelty trajectories. The small number of pre-1840 books in PG19 (fewer than 29 per decade) limits our ability to extend this analysis further back, but the 80-year trend is statistically robust.

\begin{table}[t]
\centering
\caption{Narrative shape metrics by publication decade, 1840--1910.}
\label{tab:era}
\small
\begin{tabular}{@{}lrrrrr@{}}
\toprule
Decade & $N$ & T/I & \%Green & \%Blue & \%Red \\
\midrule
1840s & 29 & 1.078 & 24.1 & 24.1 & 51.7 \\
1850s & 50 & 1.113 & 12.0 & 24.0 & 64.0 \\
1860s & 57 & 1.101 & 12.3 & 29.8 & 57.9 \\
1870s & 49 & 1.111 & 8.2 & 46.9 & 44.9 \\
1880s & 70 & 1.087 & 14.3 & 34.3 & 51.4 \\
1890s & 157 & 1.042 & 11.5 & 49.0 & 39.5 \\
1900s & 210 & 1.032 & 20.0 & 40.5 & 39.5 \\
1910s & 272 & 1.024 & 8.5 & 58.1 & 33.5 \\
\bottomrule
\end{tabular}
\end{table}


\section{Discussion}
\label{sec:discussion}

\subsection{Semantic Novelty as a Complementary Lens}

Our findings demonstrate that semantic novelty captures a dimension of narrative structure distinct from sentiment arcs. Where \citet{reagan2016} found six emotional arcs, our eight clusters describe information-density trajectories. These taxonomies are complementary, not competing: a ``man in a hole'' sentiment arc (negative dip followed by recovery) could accompany any of our eight novelty profiles. A picaresque novel might pair that arc with a high-volume novelty curve as the protagonist encounters successive strange environments; a psychological drama might pair it with a plateau profile as the emotional descent is explored through deepening elaboration of a constrained semantic field. These two frameworks capture different structural properties---affect versus information density---and future work integrating both into a joint feature space is a natural extension.

The strength of volume as a length-independent readership predictor (partial $\rho = 0.32$) suggests that the information-theoretic dimension---specifically, how much semantic territory a text traverses---constitutes an important predictor of reading behavior. Books that range widely across ideas, settings, and registers attract more readers, independent of their page count. This aligns with \citet{berlyne1971}'s foundational work on aesthetic arousal, which posits that moderate levels of novelty and complexity maximize hedonic value, and with \citet{loewenstein1994}'s information-gap theory of curiosity, in which engagement is sustained by the perceived gap between what one knows and what one wants to know.

\subsection{Comparison with Toubia et al.}

Our analysis of \citeauthor{toubia2021}'s (\citeyear{toubia2021}) metrics, after controlling for the book-length confound (Section~\ref{sec:confound}), reveals both convergences and divergences with their television findings. Speed becomes a significant positive predictor after length control (partial $\rho = 0.19$), consistent with Toubia's finding that faster narratives are preferred. The volume reversal is the most striking difference: Toubia found higher volume predicted \emph{lower} TV ratings, while we find it predicts \emph{higher} book downloads (partial $\rho = 0.32$). This may reflect the longer format: in a 30-minute episode, high semantic variance can feel chaotic, but in a book-length text, sustained exploration of diverse semantic territory reads as richness rather than incoherence.

Circuitousness, which Toubia did not test as a quality predictor, shows a dramatic confound: its raw $\rho = 0.41$ drops to partial $\rho = 0.11$ after length control. This cautionary example illustrates that summative path metrics computed over variable-length sequences can produce spuriously strong correlations in corpus studies where text length itself predicts the outcome variable.

\subsection{Genre as Information Contract}

The extremely strong genre--shape association ($\chi^2 = 2121.6$, $p < 10^{-242}$) suggests that genre conventions function as implicit information-delivery contracts---cognitive schemas \citep{swales1990} that shape reading expectations at a structural level. Fiction's plateau dominance (65\% blue) indicates that novels establish their semantic world early and maintain consistent novelty thereafter---readers expect ongoing engagement without radical departures. Non-fiction's higher T/I ratios and lower volume suggest a more directed information-delivery model: front-load key concepts, then elaborate within a narrower semantic band. Poetry's distinctive profile---lowest volume and highest convergence rate (20\% green)---reflects semantic concentration: every word carries weight, and the semantic territory is compressed rather than expansive.

These genre-specific profiles have practical implications for authors and publishers. A novel with a Science-like novelty profile (high initial novelty, steep descent) may feel ``too informational'' to fiction readers, while a history text with a Fiction-like profile (steady plateau) may feel insufficiently structured. Deviating substantially from genre-typical novelty profiles may mark a text as generically anomalous, potentially affecting reception.

\subsection{Historical Evolution}

The trend toward increasing predictability (1840--1910) maps intriguingly onto literary-historical period boundaries and the transformation of the reading public documented by \citet{altick1957}. The mid-Victorian period (1840s--1860s), characterized by experimental novels and diverse forms, shows higher T/I ratios and more divergent curves. The late Victorian period (1870s--1890s), marked by professionalization of publishing and the rise of circulating libraries such as Mudie's \citep{griest1970}, shows a transition. As \citet{griest1970} documents, lending library proprietors actively curated their collections and influenced what publishers produced, potentially selecting for more conventionally structured narratives. The Edwardian period (1900s--1910s), with its emphasis on realist fiction and established publishing houses, shows the most predictable profiles. This alignment between quantitative narrative shape metrics and qualitative book history provides mutual validation and suggests that corpus-scale narrative analysis can serve as a quantitative complement to traditional literary history.

\subsection{DTW vs.\ Euclidean Clustering}

Our comparison of DTW-based and Euclidean-based clustering yields a notable finding: despite low agreement between the two methods (ARI = 0.11), Euclidean clustering better predicts downloads (Kruskal-Wallis $H = 127.6$ vs.\ $H = 55.2$). This suggests that readers' experience is better modeled by fixed-alignment comparison---what happens at a given point in the narrative---rather than time-warped matching. DTW's invariance to temporal pacing, while valuable for abstract shape recognition, obscures exactly the positional information that may matter for reader engagement. Readers encounter narratives sequentially; a climax that arrives at chapter two versus chapter twenty creates a different experiential structure even if the abstract shape is the same. Fixed-alignment comparison preserves this positional sensitivity, which may explain its superior predictive performance.

\subsection{Limitations}

Several limitations constrain our findings. First, the PG19 corpus is restricted to pre-1920 English literature, and results may not generalize to modern or non-English texts. Second, download count is an imperfect proxy for readership quality, reflecting accessibility, fame, and curriculum assignments as much as intrinsic appeal; it measures acquisition rather than consumption or engagement. Gutenberg downloads are further skewed by public-domain availability (popular authors are repackaged across platforms) and institutional bulk downloads. Our within-genre robustness checks (Section~\ref{sec:volume}) partially address this concern: the volume effect is strongest within Fiction and Drama, where downloads more plausibly reflect reading interest, and weakest in reference-oriented genres like History and Science. Third, SBERT embeddings with \texttt{all-mpnet-base-v2} capture a specific notion of semantic similarity; replication with alternative encoders (Universal Sentence Encoder, multilingual SBERT, or larger models) would strengthen the findings, particularly for the length-independent metrics where model dependence is the primary remaining concern. Fourth, paragraph boundaries are heuristically determined and may not align with meaningful discourse units. Fifth, our genre classification is regex-based and approximate; some books are misclassified or forced into a single category when multiple genres apply. Sixth, the length confound analysis (Section~\ref{sec:confound}) demonstrates that several initially promising metrics are largely length proxies, and multivariate controls---while informative---cannot fully resolve confounds from unobserved variables such as author prestige or canonical status.

Despite these limitations, the consistency of our findings across multiple analytical approaches (clustering, partial correlations, multivariate regression, shapelet discovery, SAX analysis), the very large sample size, and the convergence with prior theoretical frameworks \citep{schmidhuber2009, toubia2021} provide confidence in the core results---particularly the length-independent finding that semantic volume predicts readership.


\section{Conclusion}
\label{sec:conclusion}

We have introduced semantic novelty---the cosine distance between each paragraph's embedding and the running centroid of all preceding paragraphs---as a new information-theoretic measure for characterizing narrative structure. Applied to 28,606 books from the PG19 corpus, this measure reveals:

\begin{enumerate}
    \item \textbf{Eight canonical narrative shape archetypes}, from Steep Descent to Steep Ascent, providing a finer-grained taxonomy than previous classification systems.
    \item \textbf{Volume as the strongest length-independent predictor of readership} (partial $\rho = 0.32$): books that traverse more semantic territory attract more readers, independent of page count. The apparently stronger raw circuitousness correlation ($\rho = 0.41$) is largely a book-length confound ($\rho_{\text{circ,length}} = 0.93$), illustrating a methodological hazard for corpus-scale narrative studies.
    \item \textbf{Strong genre--shape constraints} ($p < 10^{-242}$), revealing that genres encode implicit information-delivery contracts.
    \item \textbf{Historical trends} toward increasing predictability (1840--1910), coinciding with the professionalization of publishing.
    \item \textbf{High symbolic uniqueness} (85\% in SAX representation), indicating that each book traces a nearly unique semantic trajectory despite clustering into broad families.
\end{enumerate}

These findings complement existing work on sentiment arcs and topic trajectories by demonstrating that information density---specifically, the breadth of semantic territory explored---constitutes an important, independently predictive dimension of narrative structure.

Future work should extend this analysis to modern corpora (e.g., BookCorpus \citep{zhu2015}), non-English literatures using multilingual sentence embeddings \citep{reimers2020multilingual}, and non-literary text types (academic papers, journalism, podcasts). Reader studies pairing semantic novelty curves with physiological measures (eye tracking, skin conductance, EEG) could validate whether the computational measure corresponds to subjective experiences of surprise and engagement. Perhaps most ambitiously, semantic novelty profiles could guide text generation, enabling authors or AI systems to ``steer'' narratives toward target information-density trajectories---a practical application in AI-assisted creative writing.

The complete dataset---28,606 books with full novelty curves, PAA vectors, cluster assignments, Toubia metrics, and SAX signatures---is publicly available at \url{https://huggingface.co/datasets/wfzimmerman/pg19-semantic-novelty} under a CC-BY-4.0 license.


\section*{Acknowledgments}

This work uses the PG19 corpus \citep{rao2023pg19} derived from Project Gutenberg. Sentence embeddings were computed using SBERT \citep{reimers2019}. We thank the open-source contributors to sentence-transformers, scikit-learn, and the broader Python scientific computing ecosystem. All analyses were conducted using open-source tools.



\appendix

\section{Cluster Centroids}
\label{app:centroids}

Table~\ref{tab:centroids} provides the 16-segment PAA centroid vectors for each of the eight narrative shape clusters, enabling direct reproduction of our classification. All values are z-scored (mean 0, standard deviation 1 within each book before aggregation).

\begin{table*}[t]
\centering
\caption{Ward-linkage cluster centroids (16-segment PAA, z-scored). Each row represents the centroid of one cluster.}
\label{tab:centroids}
\footnotesize
\setlength{\tabcolsep}{2.5pt}
\begin{tabular}{@{}l*{16}{r}@{}}
\toprule
Cluster & $s_1$ & $s_2$ & $s_3$ & $s_4$ & $s_5$ & $s_6$ & $s_7$ & $s_8$ & $s_9$ & $s_{10}$ & $s_{11}$ & $s_{12}$ & $s_{13}$ & $s_{14}$ & $s_{15}$ & $s_{16}$ \\
\midrule
Steep Desc.     & .55 & .69 & .50 & .26 & .11 & $-.01$ & $-.12$ & $-.16$ & $-.23$ & $-.24$ & $-.27$ & $-.27$ & $-.27$ & $-.26$ & $-.23$ & $-.02$ \\
Grad.\ Desc.   & $-2.25$ & 1.36 & .78 & .25 & .21 & .04 & $-.02$ & $-.13$ & $-.16$ & $-.17$ & $-.18$ & $-.19$ & $-.12$ & $-.17$ & $-.16$ & .14 \\
Early Plat.     & .55 & .09 & $-.06$ & $-.10$ & $-.11$ & $-.11$ & $-.10$ & $-.10$ & $-.08$ & $-.04$ & .01 & $-.01$ & .00 & .02 & $-.03$ & .09 \\
Late Plat.      & $-.16$ & $-.28$ & $-.24$ & $-.18$ & $-.10$ & $-.03$ & .03 & .09 & .12 & .11 & .09 & .05 & .07 & .08 & .10 & .24 \\
U-Shape         & .32 & .10 & $-.04$ & $-.09$ & $-.13$ & $-.16$ & $-.15$ & $-.16$ & $-.16$ & $-.17$ & $-.16$ & $-.18$ & $-.16$ & $-.14$ & .06 & 1.21 \\
Flat            & $-.00$ & .02 & .08 & .10 & .08 & .06 & .04 & .02 & $-.01$ & $-.03$ & $-.05$ & $-.05$ & $-.06$ & $-.09$ & $-.09$ & $-.03$ \\
Grad.\ Asc.    & .03 & $-.18$ & $-.21$ & $-.20$ & $-.22$ & $-.21$ & $-.22$ & $-.23$ & $-.24$ & $-.26$ & $-.25$ & $-.21$ & .01 & .47 & .93 & .98 \\
Steep Asc.      & $-.13$ & $-.36$ & $-.37$ & $-.40$ & $-.39$ & $-.37$ & $-.36$ & $-.30$ & $-.18$ & .03 & .29 & .54 & .59 & .51 & .43 & .47 \\
\bottomrule
\end{tabular}
\end{table*}

\section{Genre Classification Rules}
\label{app:genres}

Genre classification uses regex matching against Library of Congress subject headings in the following priority order: Fiction (\texttt{fiction|novel|stories|tales|romance|fantasy}), Poetry (\texttt{poetry|poems|verse|sonnet|ballad}), Drama (\texttt{drama|play|theater|comedy|tragedy}), History (\texttt{history|historical|war|battle|military}), Science (\texttt{science|scientific|natural history|zoolog|botan}), Philosophy/Religion (\texttt{philosoph|religio|theolog|bible|christian}), Travel/Geography (\texttt{travel|voyage|explor|geograph}), Biography (\texttt{biograph|correspondence|diaries|letters|memoir}), Children's/Juvenile (\texttt{juvenile|children|fairy tale}), Social Science (\texttt{political|economi|social|law|government}). Books matching none of these are classified as Other.

\section{Cluster Selection Diagnostics}
\label{app:kselection}

Table~\ref{tab:wcss} reports within-cluster sum of squares (WCSS) and mean silhouette scores for $k = 2$ through $12$ on the 8,000-book Ward-linkage sample. Silhouette peaks at $k=2$ (0.091) and declines monotonically, indicating that the data form a continuum rather than discrete clusters. WCSS shows an elbow around $k=5$--$6$, with marginal reduction dropping below 2\% after $k=7$. We select $k=8$ as a pragmatic balance between granularity and interpretability.

\begin{table}[h]
\centering
\caption{Cluster selection diagnostics ($n = 8{,}000$ subsample).}
\label{tab:wcss}
\small
\begin{tabular}{@{}rrrr@{}}
\toprule
$k$ & WCSS & Silhouette & Marginal \% \\
\midrule
2 & 112,325 & 0.091 & --- \\
3 & 105,243 & 0.059 & 6.3\% \\
4 & 101,064 & 0.058 & 4.0\% \\
5 & 97,663 & 0.029 & 3.4\% \\
6 & 95,643 & 0.029 & 2.1\% \\
7 & 93,789 & 0.022 & 1.9\% \\
\textbf{8} & \textbf{92,147} & \textbf{0.020} & \textbf{1.8\%} \\
9 & 90,579 & 0.021 & 1.7\% \\
10 & 89,540 & 0.018 & 1.1\% \\
11 & 88,507 & 0.005 & 1.2\% \\
12 & 87,541 & 0.005 & 1.1\% \\
\bottomrule
\end{tabular}
\end{table}

\section{Dataset Schema}
\label{app:schema}

The publicly available dataset contains 24 columns per book: \texttt{gutenberg\_id} (int), \texttt{title} (string), \texttt{authors} (list[string]), \texttt{pub\_year} (int), \texttt{subjects} (list[string]), \texttt{bookshelves} (list[string]), \texttt{download\_count} (int), \texttt{primary\_genre} (string), \texttt{paragraph\_count} (int), \texttt{mean\_novelty} (float), \texttt{std\_novelty} (float), \texttt{ti\_ratio} (float), \texttt{trend\_slope} (float), \texttt{mean\_compression\_progress} (float), \texttt{curve\_type\_3} (string), \texttt{cluster\_8} (int), \texttt{cluster\_name} (string), \texttt{speed} (float), \texttt{volume} (float), \texttt{circuitousness} (float), \texttt{reversal\_count} (int), \texttt{sax\_16\_5} (string), \texttt{novelty\_curve} (list[float]), \texttt{paa\_16} (list[float]).

\end{document}